\documentclass[runningheads]{llncs}

\pdfoutput=1

\usepackage{eccv}

\usepackage{eccvabbrv}

\usepackage{graphicx}
\usepackage{booktabs}

\usepackage{multirow}
\usepackage{multicol}
\usepackage{wrapfig}
\usepackage{placeins}

\usepackage[accsupp]{axessibility}  

\usepackage{hyperref}

\usepackage{orcidlink}

\begin{document}

\title{Task-driven Processing with Coarse-to-Fine Glimpse-based Active Perception} 


\author{Oleh Kolner\inst{1,2} \and
Thomas Ortner\inst{1} \and
Stanisław Woźniak\inst{1} \and
Angeliki Pantazi\inst{1}
}

\authorrunning{O.~Kolner et al.}

\institute{IBM Research, Zurich, Switzerland\\ 
\and Graz University of Technology, Graz, Austria \\
\email{olk@zurich.ibm.com}}

\maketitle

\begin{abstract}
    State-of-the-art vision models process images in their entirety, lacking the ability to selectively zoom in on relevant regions.
    This limitation is particularly acute in scenarios where processing must be conditioned on a specific task -- such as instance detection, which requires localizing a specific object in a high-resolution, cluttered scene.
    In such settings, critical details are easily lost as images are often resized to match the model dimensions and computational constraints. 
    We introduce Coarse-to-Fine Glimpse-based Active Perception (CF-GAP), a task-driven front-end that enhances high-resolution processing of existing instance detectors. CF-GAP selectively directs a sequence of limited-view glimpses across the scene, utilizing task information to iteratively refine focus on the most relevant regions.
    These localized regions are then processed at high resolution by a downstream instance detector.
    By avoiding full-image processing and eliminating irrelevant confounding information, CF-GAP improves Average Precision (AP) by up to 20\% across various state-of-the-art instance detectors on the HR-InsDet and Robotools benchmarks, while further enabling lightweight detectors to outperform their larger counterparts.

    \keywords{Active perception \and Bio-inspired vision \and Instance detection} 
\end{abstract}

\newcommand{\bSTT}{OTS-FM$_{\text{STT}}$}
\newcommand{\bSAM}{OTS-FM$_{\text{SAM}}$}
\newcommand{\bMobileSAM}{OTS-FM$_{\text{MobileSAM}}$}
\newcommand{\bGDINO}{OTS-FM$_{\text{GroundingDINO}}$}


\newcommand{\MainArchitecture}[1]{
    \begin{figure}[#1]
        \centering
        \includegraphics[width=\linewidth]{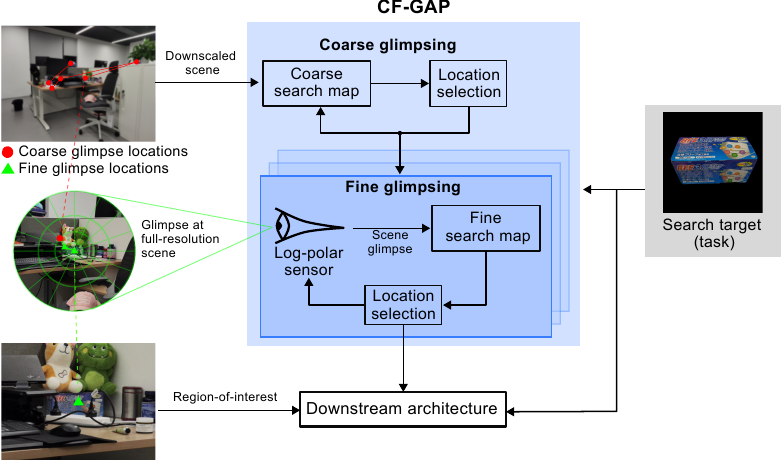}
        \caption{
            Coarse-to-Fine Glimpse-based Active Perception (CF-GAP) iteratively directs a series of glimpses across a high-resolution scene, using task-driven search maps to progressively narrow focus onto the likely search target. The resulting regions of interest are passed at high resolution to the downstream architecture for the final detection.
        }
        \label{fig:OverallArch}
    \end{figure}
}

\newcommand{\FineSearchMap}[1]{
    \begin{figure}[#1]
        \centering

        \includegraphics[width=0.85\linewidth]{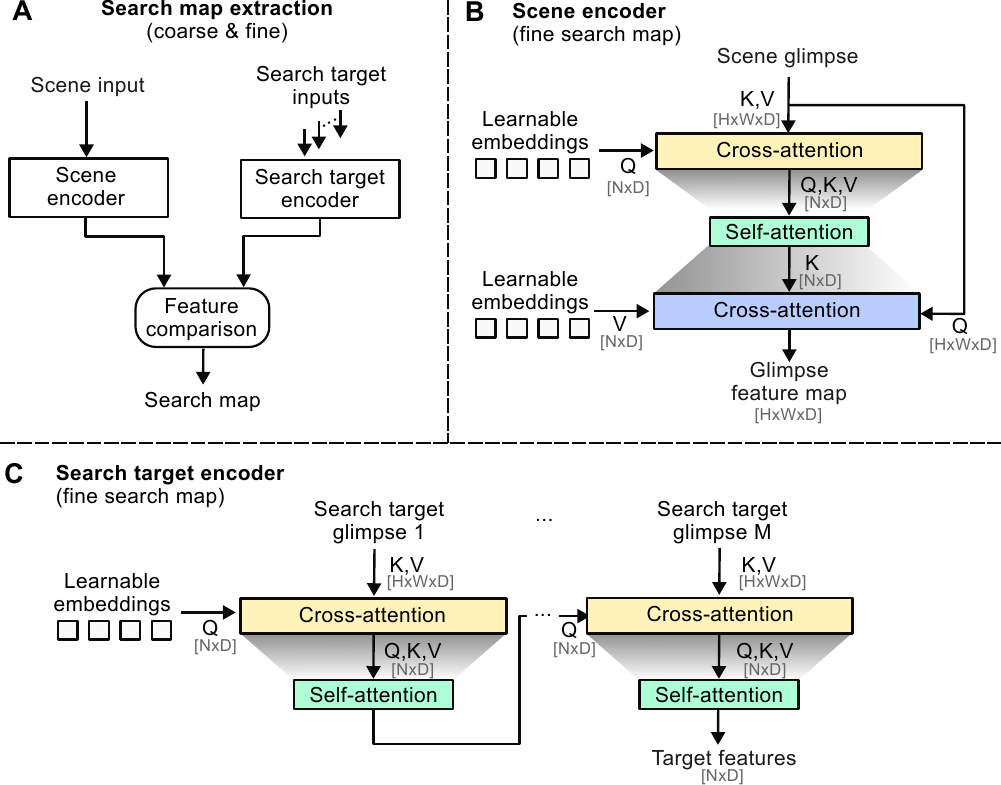}

        \caption{
            \textbf{(A)} General architecture for extracting coarse and fine search maps. 
            \textbf{(B)} Scene encoder compresses the scene glimpse into a compact set of latent embeddings via cross-attention, where a small number of learnable embeddings attend to the full spatial input ($N \ll H\times W$).  
            The compressed representation is then processed through self-attention. 
            A second cross-attention decodes the resulting representation back to the original spatial dimensions, where a distinct set of learnable value embeddings defines the output feature space. Q, K, and V denote queries, keys, and values of each attention block.
            \textbf{(C)} Search target encoder shares the cross- and self-attention blocks with the scene encoder (marked by the same color) and iteratively attends to multiple search target glimpses to produce compact target features to be compared with the glimpse feature map.
        }
        \label{fig:FineSearchMap}
    \end{figure}
}

\newcommand{\LogPolarQualitative}[1]{
    \begin{figure}[#1]
        \centering
        \includegraphics[width=\linewidth]{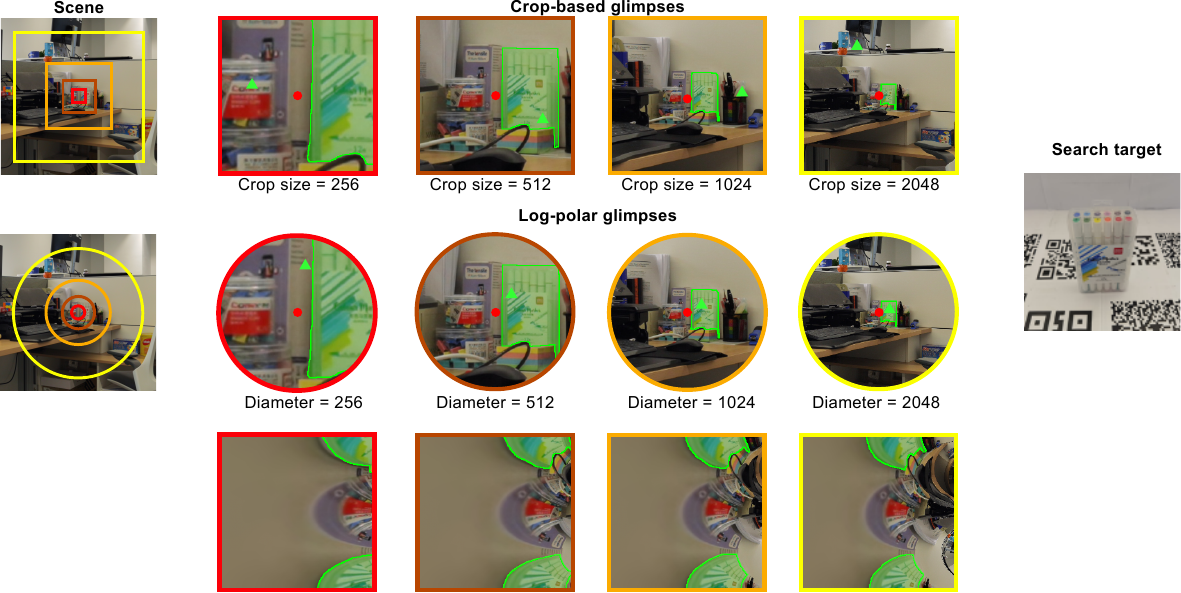}
        \caption{
        Qualitative comparison between crop-based and log-polar glimpses across different sizes. Red dots mark the initial coarse glimpse location, offset from the search target, and green triangles mark the subsequent fine glimpse location based on fine search maps extracted from each glimpse. Unlike crop-based glimpses (top row), which change drastically with size, log-polar glimpses (bottom row) remain visually stable across different diameters, as peripheral differences are compressed into the far-right part of each log-polar image. The middle row shows the regions covered by the log-polar glimpses. 
        }
        \label{fig:LogPolarQualitative}
    \end{figure}
}

\newcommand{\MainResult}[1]{
    \begin{figure*}[#1]
        \centering
        \includegraphics[width=\linewidth]{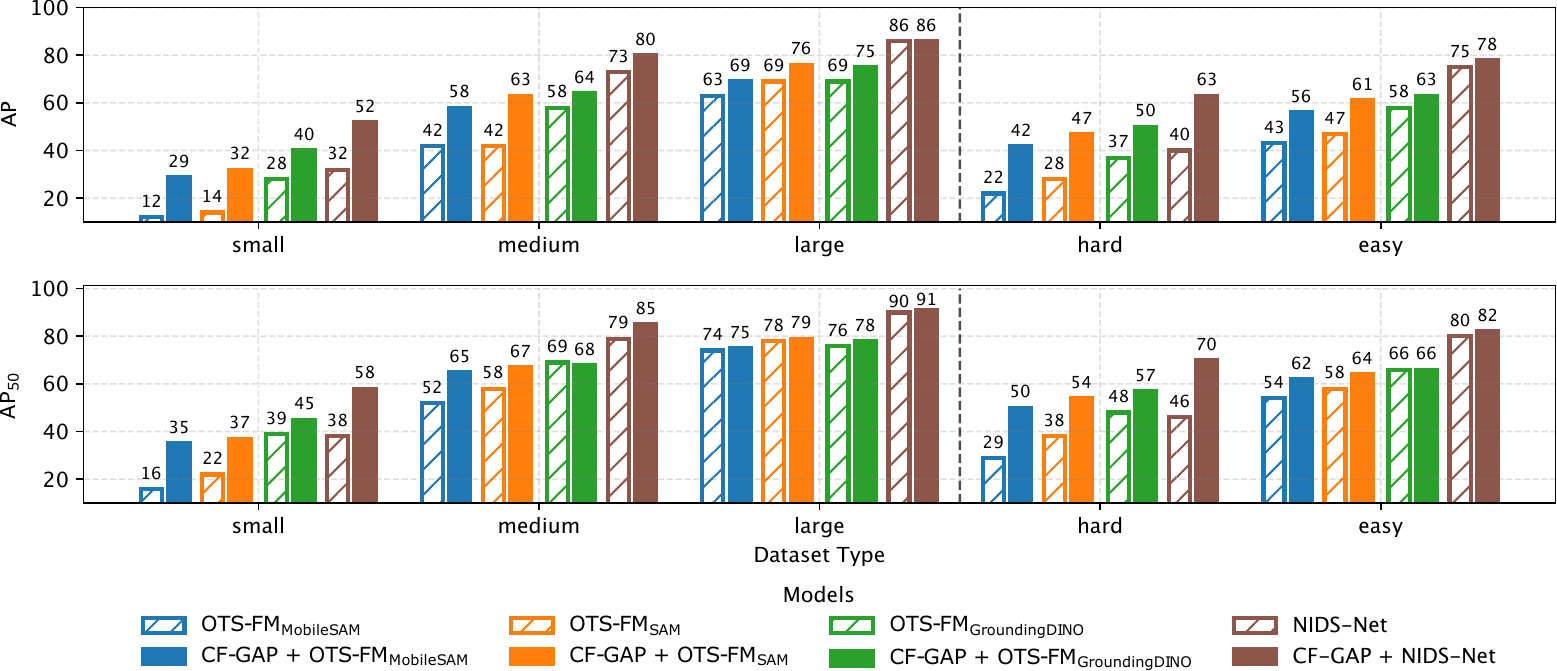}
        \caption{
            Performance on HR-InsDet dataset across two dataset groupings -- by search target sizes, and by scene types.
        }
        \label{fig:MainResult}
    \end{figure*}
}

\newcommand{\MainResultRobotools}[1]{
    \begin{figure}[#1]
        \centering
        \includegraphics[width=0.9\linewidth]{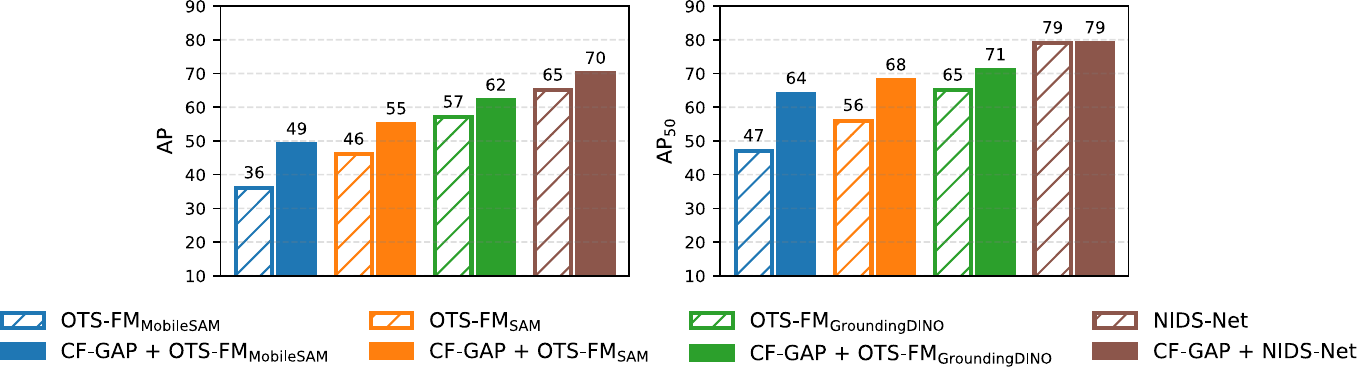}
        \caption{
            Performance on Robotools dataset, testing generalization to novel objects unseen during training.
            Baseline results in dashed bars are taken from \cite{shen2025solving, nidsnet}
        }
        \label{fig:MainResultRobotools}
    \end{figure}
}

\newcommand{\QualitativeGlimpses}[1]{
    \begin{figure*}[#1]
        \centering
        \includegraphics[width=1\linewidth]{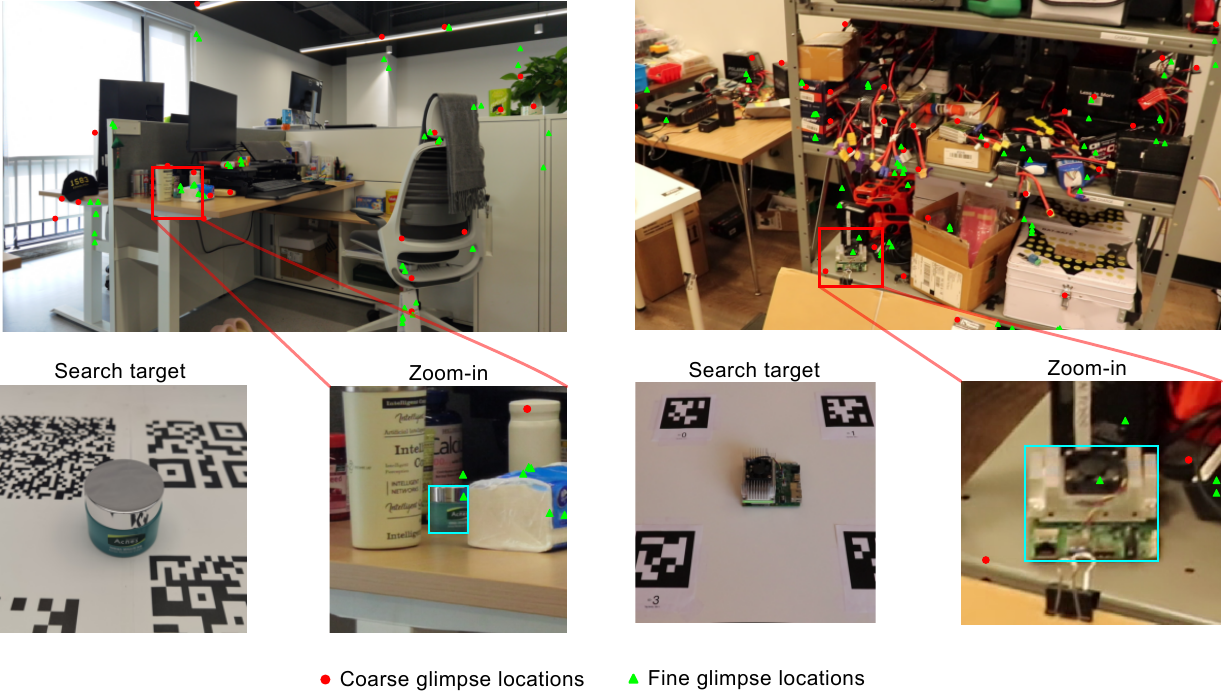}
        \caption{
            Coarse and fine glimpse locations for two scenes where baseline models failed to detect the search target (left: HR-InsDet, right: Robotools). Glimpse locations concentrate on a small fraction of the scene rather than spreading uniformly, and fine glimpse locations visibly correct the spatial imprecision of coarse ones.
        }
        \label{fig:QualitativeGlimpses}
    \end{figure*}
}

\newcommand{\EfficiencyAnalysis}[1]{
    \begin{figure}[#1]
        \centering
        \includegraphics[width=\linewidth]{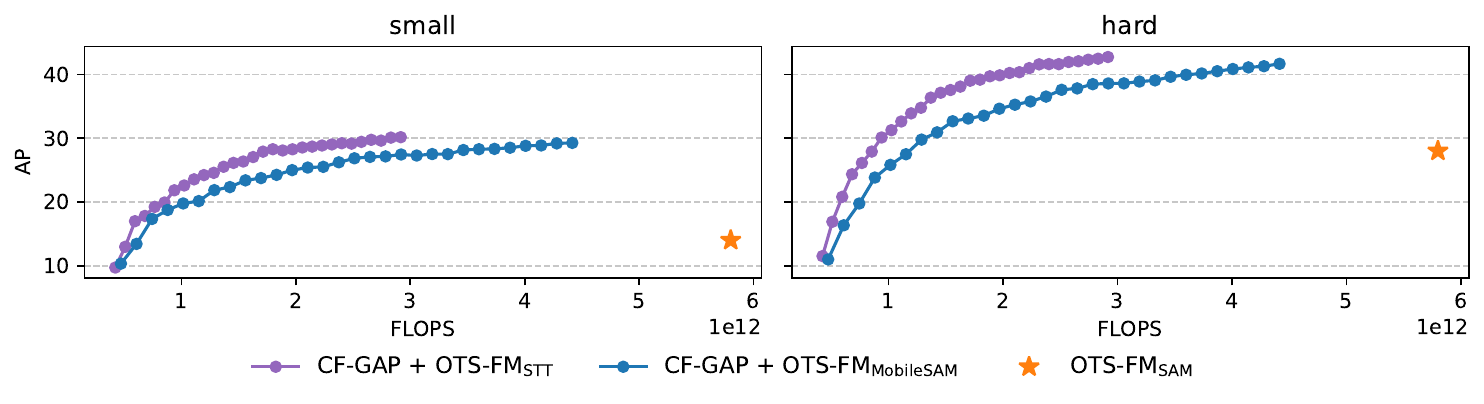}
        \caption{
            CF-GAP with efficient \bSAM~variants evaluated on the two most challenging HR-InsDet subsets. Each dot corresponds to a specific number of coarse glimpses (1 to 30). Other baselines are omitted as no efficient variants are available.
        }
        \label{fig:EfficiencyAnalysis}
    \end{figure}
}

\newcommand{\GlimpseAnalysis}[1]{
    \begin{figure}[#1]
        \centering
        \includegraphics[width=\linewidth]{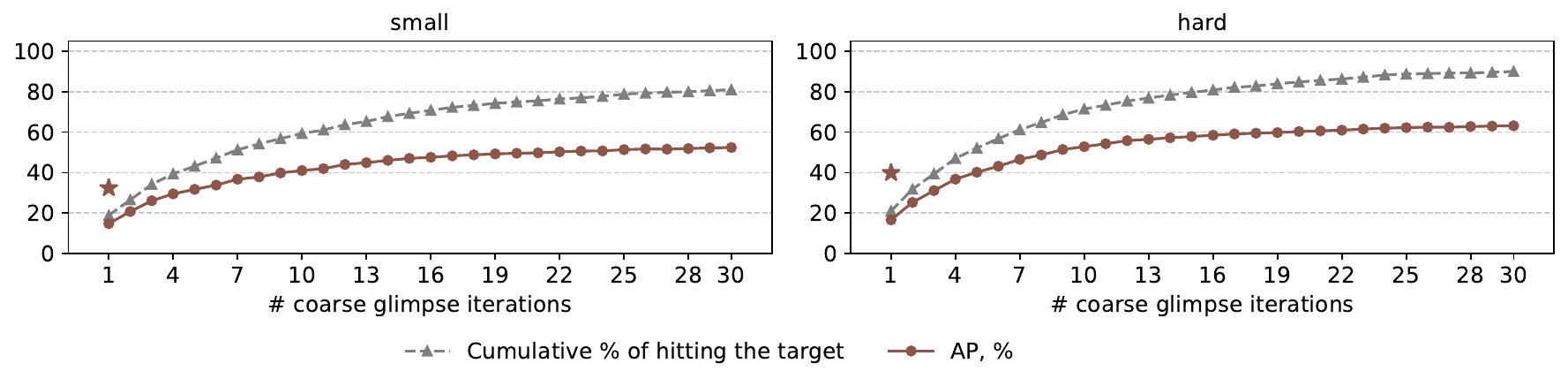}
        \caption{ 
            Impact of the number of coarse glimpses on performance. Results are shown for CF-GAP+NIDS-Net evaluated on the two most challenging HR-InsDet subsets. The stars mark the performance of standalone NIDS-Net. 
        }
        \label{fig:GlimpseAnalysis}
    \end{figure}
}

\newcommand{\LogPolarVSCropPerformance}[1]{
    \begin{figure}[#1]
        \centering
        \includegraphics[width=\linewidth]{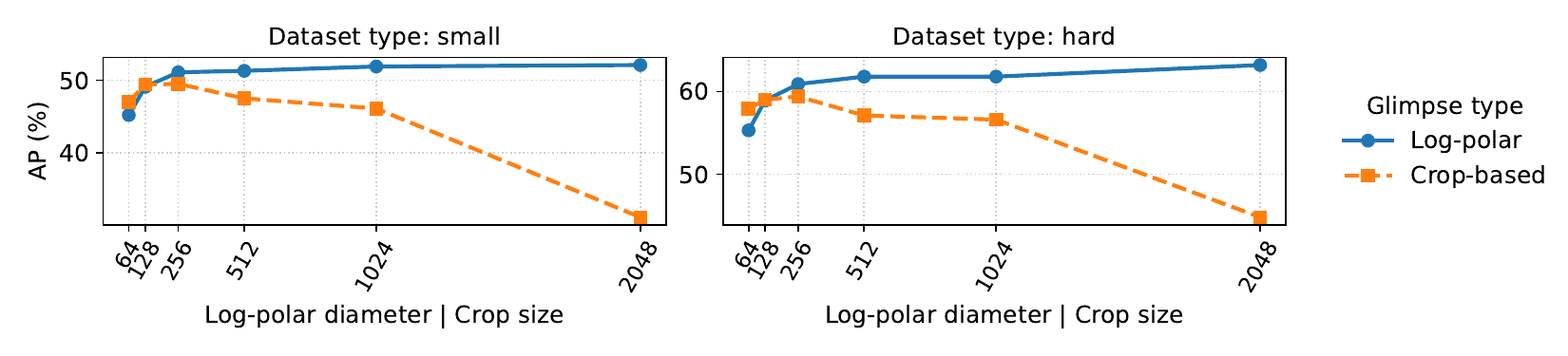}
        \caption{ 
            Performance comparison for different glimpse types and their sizes. 
            Results are for CF-GAP+NIDS-Net evaluated on the two most challenging HR-InsDet subsets.
        }
        \label{fig:LogPolarVSCropPerformance}
    \end{figure}
}


\newcommand{\FineGapAblationTable}[1]{

    \begin{table}[#1]
    \caption{Ablation of fine glimpsing. Results are shown for CF-GAP+NIDS-Net evaluated on HR-InsDet. }
    \label{tab:FineGapAblationTable}
    \centering
    \begingroup
    \fontsize{8}{9}\selectfont
    \begin{tabular}{l|ccccc}
    \multirow{2}{*}{\begin{tabular}[c]{@{}c@{}}Fine \\glimpsing\end{tabular}}  & \multicolumn{5}{c}{AP}       \\
                                             & small &medium & large & easy & hard  \\ 
    \hline
    \multicolumn{1}{c|}{no}  & 44.6  & 77.3   & 85.5  & 75.6 & 56.2    \\
    \multicolumn{1}{c|}{yes} & \textbf{52.1}  & \textbf{79.7}   & \textbf{86.1}  & \textbf{78.2} & ~\textbf{63.2 }
    \end{tabular}
    \endgroup
    \end{table}
}

\newcommand{\LocationForDA}[1]{
    
    \begin{table}[#1]
    \caption{Ablation of inputs provided to the downstream architecture. Results are shown for CF-GAP$+$NIDS-Net evaluated on HR-InsDet.}
    \label{tab:LocationForDA}
    \centering
    \begingroup
    \fontsize{8}{9}\selectfont
    \begin{tabular}{c|ccccc}
    \multirow{2}{*}{\begin{tabular}[c]{@{}c@{}}Downstream \\architecture input\end{tabular}} & \multicolumn{5}{c}{AP}                \\
                                                                                             & small & medium & large & easy & hard  \\ 
    \hline
    w/o locations                                                                    & 41.8  & 73.1   & 86.0  & 74.5 & 49.5  \\
    with locations                                                                   & \textbf{52.1}  & \textbf{79.7}   & \textbf{86.1}  & \textbf{78.2} & ~\textbf{63.2 }
    \end{tabular}
    \endgroup
    \end{table}
}

\newcommand{\PatchAblationTable}[1]{%
    \begin{table}[#1]
    \centering
    \caption{
    Comparison to patched baselines that split the scene into smaller, separately processed patches. Results on the two most challenging HR-InsDet subsets.
    }
    \label{tab:PatchAblationTable}
    \begingroup
    \fontsize{8}{9}\selectfont

    \begin{tabular}{l|cc}
    \multirow{2}{*}{Model} & \multicolumn{2}{c}{AP} \\ 
                                    & small & hard  \\ 
    \hline
    \bMobileSAM                       & 12.4    & 22.0    \\
    Patched \bMobileSAM           & 18.4    & 23.2    \\
    CF-GAP + \bMobileSAM                   & \textbf{29.3}    & \textbf{41.7}    \\ 
    \hline
    \bSAM                             & 14.6    & 28.0    \\
    Patched \bSAM                 & 18.9    & 24.2    \\
    CF-GAP + \bSAM                         & \textbf{32.4}    & \textbf{47.1}    \\ 
    \hline
    \bGDINO                   & 28.8    & 37.2    \\
    Patched \bGDINO       & 22.4    & 28.1    \\
    CF-GAP + \bGDINO               & \textbf{39.6}    & \textbf{50.2    }\\

    \hline
    NIDS-Net                   & 32.4    & 39.9    \\
    Patched NIDS-Net       & 50.1    & 49.1    \\
    CF-GAP + NIDS-Net               & \textbf{52.1}    & \textbf{63.2}
    \end{tabular}
    \endgroup

    \end{table}
}

\newcommand{\MissErrorAnalysisTable}[1]{
    \begin{table}[#1]
    
    \caption{Percentage of scenes in the two most challenging HR-InsDet subsets, where the search target was found, but not necessarily correctly recognized.
    }
    \label{table:MissErrorAnalysisTable}
    \centering
    \begingroup
    \fontsize{8}{9}\selectfont
    \begin{tabular}{l|cc}
    \multicolumn{1}{c|}{\multirow{2}{*}{Model}} & \multicolumn{2}{c}{Target found, \%}  \\
    \multicolumn{1}{c|}{}                       & small & hard                               \\ 
    \hline
    \bSAM                                         & 42.1    & 56.3                                 \\
    NIDS-Net                               & 73.3    & 75.2                                \\
    \hline
    CF-GAP~                                     & \textbf{81.9}    & \textbf{89.6}                                 \\
    CF-GAP~ w/o fine glimpsing                  & 69.5    & 74.6                                 \\
    \end{tabular}
    \endgroup
    \end{table}
}

\newcommand{\FlopsReport}[1]{
    \begin{table}[#1]
        \caption{Cost breakdown for CF-GAP paired with downstream architectures. $N_c$ and $N_f$ correspond to the number of coarse and fine glimpses, respectively.}
        \label{tab:flops_report}
        \centering
        \resizebox{0.7\columnwidth}{!}{%
        \begin{tabular}{ll}
        \toprule
        Component & GFLOPs \\
        \midrule
        Coarse search map, SM$_c$ (MobileNet) & 60 \quad (per scene) \\
        Fine search map,   SM$_f$ (Perceiver-based encoders) & 1 \quad (per glimpse) \\
        \midrule
        \textbf{CF-GAP internal cost:} $\mathrm{SM}_c + N_c{\times}N_f{\times}\mathrm{SM}_f$
          & $60 + N_c{\times}N_f{\times}1$ \\
        \midrule
        Downstream architecture (DA) & \\
        \{OTS$_\text{STT}$\,$|$\,OTS$_\text{MobileSAM}$\,$|$\,OTS$_\text{SAM}$\,$|$\,NIDS-Net\}
          & \{80\,$|$\,130\,$|$\,5800\,$|$\,800\} \\
        \midrule
        \textbf{Total cost:} $[ \text{CF-GAP internal cost} ]$ + $N_c\times [\text{DA cost}]$ \\
        \bottomrule
        \end{tabular}
        }
    \end{table}
}


\section{Introduction}\label{sec:Intro}
\MainArchitecture{t}

State-of-the-art vision models process entire images uniformly, lacking the ability to selectively zoom into task-relevant regions for detailed analysis. Moreover, they require images to be resized to the fixed input dimensions used during pre-training, often obscuring small but important regions. These limitations hinder performance on high-resolution images where task-relevant content is small or cluttered \cite{v_star, zhang2025mllms, hr_ins_det}. A further shortcoming is that task-specific information (\eg, a textual prompt or a specific object to be localized) is typically incorporated only after a costly feature extraction stage. However, early integration of such information could significantly reduce the image area requiring intensive computation and mitigate the influence of irrelevant or distracting features.

Human vision, by contrast, is inherently task-driven: we actively seek visual information based on what we aim to accomplish, rather than passively processing every detail \cite{Yarbus1967, hayhoe2005eye}. For example, in an office environment shown in \cref{fig:OverallArch}, one often needs to find a specific box or tool -- not just any object from those categories. 
This targeted search corresponds to instance detection, where the goal is to localize a specific object instance given a few visual examples, as opposed to classical object detection, which seeks to identify all objects in a scene. 
As opposed to typical vision models, humans actively leverage instance-specific features as task cues for efficient visual search \cite{wolfe2010visual}.
A further hallmark of human vision is the foveal structure of the eye, which provides non-uniform resolution -- highest at the center and decreasing toward the periphery -- balancing detailed information with a broad field of view \cite{schwartz1977spatial, freeman2011metamers}. 
Humans exploit this balance through eye movements (saccades) that follow a coarse-to-fine strategy: macrosaccades, guided by low-resolution cues, direct gaze to promising areas, which are then analyzed in detail through microsaccades \cite{navon1977forest, yu2023good, rolfs2009microsaccades}.

Inspired by saccadic processing, the recently proposed Glimpse-based Active Perception (GAP) model \cite{gap} fixates on salient -- not necessarily task-relevant -- image regions to solve synthetic visual reasoning tasks, demonstrating strong out-of-distribution generalization and, thus, the effectiveness of processing only selected image parts.
Building on this, we propose Coarse-to-Fine GAP (CF-GAP), which replaces uniform image processing by selectively directing a sequence of limited-view glimpses (\cref{fig:OverallArch}).
This sequence is generated through a nested process of coarse and fine glimpsing, guided by a task-specific object to be localized, referred to as a search target.
Coarse glimpsing relies on correlations between the search target's features and the downscaled scene to identify an initial coarse glimpse location.
There, a fovea-inspired log-polar sensor extracts a limited-view glimpse from the full-resolution scene, magnifying the visual details around the glimpse location while preserving peripheral information at diminishing resolution.
Operating on this focused view, fine glimpsing directs the log-polar sensor in a closed loop to iteratively refine the initial location toward a likely search target.
The resulting fine glimpse locations define a localized region of interest (RoI) that is passed to a downstream architecture at high-resolution to determine whether the search target is present at those locations.

Our work focuses on the instance detection problem, where visual processing must be conditioned on a specific search target. 
We design CF-GAP as a frontend module that integrates seamlessly with existing instance detectors as downstream architectures, providing them with targeted, high-resolution input. 
We combine CF-GAP with several state-of-the-art detectors and evaluate on HR-InsDet \cite{hr_ins_det} and Robotools \cite{robotools} as two challenging benchmarks.
We observe significant performance improvement across all models. Notably, integrating CF-GAP with small models optimized for edge devices allowed them to match, and in some difficult cases, even surpass the performance of their larger versions.    

To summarize, our contributions are the following:
\begin{enumerate}
\item We introduce a bio-inspired, coarse-to-fine processing scheme to selectively explore high-resolution scenes, replacing exhaustive full-scene processing.
\item We design our model as a front-end module that can be seamlessly integrated with any existing instance detector, enhancing it with targeted, high-resolution input.
\item We show that our approach significantly boosts the performance of state-of-the-art instance detectors, particularly in cluttered and complex scenes.
\end{enumerate}

\section{Related Work}  
Early instance detection methods trained CNN-based object detectors with a separate class per object instance, using the cut-paste-learn (CPL) framework~\cite{cut_paste_learn}, where objects are pasted into random backgrounds.  
State-of-the-art instance detection methods~\cite{hr_ins_det, shen2025solving, nidsnet} adopt a two-stage pipeline.
First, all candidate objects in the scene are detected using foundation models pre-trained on large-scale data, in particular Segment Anything Model (SAM)~\cite{sam, sam2} and GroundingDINO~\cite{grounding_dino}.
Second, the candidates are matched against visual examples of the search target using DINOv2~\cite{dinov2} features, with the best match returned as the final prediction.
Recent extensions improve the matching stage by fine-tuning DINOv2~\cite{shen2025solving} or training a weight adapter~\cite{nidsnet} on specific object instances.
Critically, all the aforementioned methods are from the ground up not task-driven, since the first stage of proposal detection involves exhaustive analysis of the entire image to detect all possible objects irrespective of the search target. 
Our approach leverages the information about the task of detecting a specific search target to steer detection toward only promising regions.

Multiple approaches drew high-level inspiration from saccadic processing and showed compelling results across various tasks, including image classification \cite{mnih_recurrent_2014_2, xu_et_al, elsayed2019saccader}, object detection \cite{ba2014multiple, ibrayev2024exploring, ibrayev2024toward}, visual exploration \cite{pardyl2025adaglimpse}, and visual reasoning \cite{wozniak2023visual, gap}. 
However, none of them allows for conditioning by task information (\eg by a search target) at inference to guide the image exploration. 
While our approach is built upon GAP \cite{gap}, the original framework lacks a coarse-to-fine, task-driven search strategy.

Fovea-inspired vision with non-uniform resolution was explored for guiding the visual search only in simple images \cite{akbas2017object, cheung2016emergence}.
Various methods explored fovea-like log-polar imagery and its rotational and scaling equivariance properties for image classification, object detection, and image correspondence \cite{esteves2017polar, kim2020cycnn, ebel_beyond_2019}. Another approach \cite{stt} proposed a sophisticated tokenization for vision transformers (ViTs) \cite{dosovitskiy2021an}, splitting the image into patches of various sizes resembling foveated image structure. 
While prior work used fovea-inspired imagery mainly to encode a fixed view, we use it instead to navigate the coarse-to-fine glimpsing process. 

\section{Method}\label{sec:Model} 

Conceptually, CF-GAP can be viewed as an active process of steering a virtual log-polar sensor across a scene to acquire high-quality information related to the search target (\cref{fig:OverallArch}). 
More specifically, CF-GAP directs a sequence of glimpses -- limited views of an image taken at specific locations -- to pinpoint a likely search target location. 
This location defines the center of a RoI that is passed to the downstream architecture, allowing it to isolate and verify the candidate object, without exhaustively processing the entire scene.
CF-GAP operates in two nested stages of coarse and fine glimpsing guided by distinct search maps. The search maps are 2D heatmaps that highlight regions with a high probability of containing the search target.
The search map extraction follows a common scheme from \cite{zeng2021transporter, kreiman2018}: a scene encoder and a search target encoder produce features that are compared via convolution (\cref{fig:FineSearchMap}A).
The coarse and fine glimpsing stages differ in how this scheme is instantiated, as described below, with further technical details provided in Appendix A.

\FineSearchMap{!b}

\subsection{Coarse glimpsing}
CF-GAP begins with computing a coarse search map from a downscaled version of the scene and sample images of the search target taken from different viewpoints (referred to as search target examples).
The scene and search target encoders (\cref{fig:FineSearchMap}A) are instantiated with the lightweight MobileNet-V3 \cite{mobilenet}.
The features of the search target examples are averaged across spatial dimensions into a single feature vector, which is then convolved over the scene features, producing the coarse search map.
This map drives the iterative selection of coarse glimpse locations: at each iteration, the location of the highest value is selected via the winner-takes-all (WTA) strategy, followed by inhibition-of-return (IoR) that masks its surrounding region to prevent repeated selection, similar to~\cite{itti_model_1998, gap}.
Since coarse glimpse locations are derived from low-resolution imagery, they represent only rough estimates of the search target's position. Therefore, each coarse glimpse location initiates a closer inspection with fine glimpsing.

\subsection{Fine glimpsing}
Unlike coarse glimpsing, which operates on a static downscaled scene, fine glimpsing operates on glimpses of both the scene and the search target. It is a closed-loop process: 
at each iteration, a fine search map is generated from the current scene glimpse, and its 2D centroid determines the next fine glimpse location. The log-polar sensor then moves to that location and extracts a new scene glimpse, from which the next iteration proceeds.
The sensor employs a fovea-like log-polar transformation to extract high-resolution detail near the glimpse location while preserving distant context at progressively lower resolution.
The resulting focused view makes fine search map generation robust against peripheral distractors, and we analyze its advantages over cartesian cropping in \cref{sec:Results}.

Since the fine search map extraction (\cref{fig:FineSearchMap}A) operates on log-polar glimpses, using CNN-based scene and search target encoders (as for the coarse search map) becomes problematic. 
Specifically, CNNs assume that the input image has uniform resolution, treating all regions equally regardless of their position within that image. 
By contrast, log-polar glimpses preserve fine detail near the glimpse location while compressing the periphery into much fewer pixels.
We therefore propose modules based on Perceiver~\cite{perceiver}, illustrated in \cref{fig:FineSearchMap}B-C.

Each glimpse is partitioned into non-overlapping patches projected into a $D$-dimensional space, $\boldsymbol{F}_s \in \mathbb{R}^{H \times W \times D}$ and $\{\boldsymbol{F}_m\}_{m=1}^{M}$ denote the corresponding representations of the scene glimpse and $M$ search target glimpses, respectively.
The scene encoder (\cref{fig:FineSearchMap}B) first compresses the scene glimpse into a compact set of $N$ latent embeddings via cross-attention, where a small number of learnable queries attend to the full spatial input, and then processes them through self-attention:
\begin{equation}\label{eq:scene_compress_m}
    \tilde{\boldsymbol{h}}_s = \mathcal{C}\!\left(\mathsf{Q} = \boldsymbol{e}_q,\; \mathsf{KV} = \boldsymbol{F}_s\right), \qquad
    \boldsymbol{h}_s = \mathcal{S}\!\left(\mathsf{QKV} = \tilde{\boldsymbol{h}}_s\right),
\end{equation}
where $\boldsymbol{e}_q \in \mathbb{R}^{N \times D}$ is a set of learnable embeddings and $N \ll H \times W$. The compression via cross-attention alleviates the quadratic cost of the subsequent self-attention. The self-attention, in turn, integrates fine local details near the glimpse center with the broader peripheral context.
Finally, the latent embeddings are decoded back to the original spatial dimensions through a second cross-attention, producing the glimpse feature map $\boldsymbol{F}^*_s$ as the final output:
\begin{equation}\label{eq:scene_decode_m}
    \boldsymbol{F}^*_s = \mathcal{C}_{\mathrm{dec}}\!\left(\mathsf{Q} = \boldsymbol{F}_s,\; \mathsf{K} = \boldsymbol{h}_s,\; \mathsf{V} = \boldsymbol{e}_o\right).
\end{equation}
The decoding cross-attention uses a distinct set of learnable embeddings $\boldsymbol{e}_o \in \mathbb{R}^{N \times D}$ as values to decouple the feature space used for latent compression from the one used for the output feature map. This is intended to let the output feature map be optimized specifically for comparison with features extracted from the search target.

The search target encoder (\cref{fig:FineSearchMap}C) follows the same compress-and-process pattern with shared cross- and self-attention blocks, but instead of decoding back to spatial dimensions, it iteratively attends to $M$ search target glimpses to produce a set of compact target features. At each iteration $m$, the cross-attention receives the search target glimpse $\boldsymbol{F}_m$ as keys and values and the previous output as queries, followed by self-attention:
\begin{equation}\label{eq:target_cross_m}
    \tilde{\boldsymbol{h}}_T^{(m)} = \mathcal{C}\!\left(\mathsf{Q} = \boldsymbol{h}_T^{(m-1)},\; \mathsf{KV} = \boldsymbol{F}_{m}\right), \qquad
    \boldsymbol{h}_T^{(m)} = \mathcal{S}\!\left(\mathsf{QKV} = \tilde{\boldsymbol{h}}_T^{(m)}\right),
\end{equation}
with the query at the first iteration initialized to $\boldsymbol{h}_T^{(0)} = \boldsymbol{e}_q$ and the final output $\boldsymbol{h}_T = \boldsymbol{h}_T^{(M)}, \boldsymbol{h}_T\in \mathbb{R}^{N \times D}$ used as the set of target features.
These target features are then correlated with the glimpse feature map via convolution, yielding $N$ correlation maps that are averaged into a single fine search map.

\subsection{Downstream architecture}
At the end of each fine-glimpsing loop, CF-GAP provides the downstream architecture with three inputs.
First, a fixed-size RoI cropped around the last fine glimpse location.
Second, the fine glimpse location itself, which, depending on the downstream architecture, serves either as a spatial prompt (\eg for SAM-like models) or as a bounding box filter to constrain object detection.
Third, multiple visual examples of the search target from different viewpoints, used for matching with the detected candidate object.
After a predefined number of coarse and fine glimpsing iterations, the best-matched candidate object is returned.
Importantly, CF-GAP is agnostic to the choice of downstream architecture. During evaluation, we employ several state-of-the-art detectors as described in \cref{sec:Experiments}.

\section{Experiments}\label{sec:Experiments}
\subsubsection{Instance detection} consists of individual tasks, each defined by a few visual examples of a specific object to be localized in an input scene.
Unlike classical object detection, which identifies all instances of an object category, instance detection is conditioned on a particular object instance.
We consider two benchmarking datasets.
The first, HR-InsDet~\cite{hr_ins_det}, contains 100 object instances with 24 visual examples each, and 160 high-resolution scenes spanning 14 indoor scenarios.
For training, the dataset provides 200 images with random backgrounds to synthesize training data via the cut-paste-learn strategy~\cite{cut_paste_learn}, where search targets are resized and pasted onto arbitrary backgrounds.
The dataset is split into subsets by the level of clutter and occlusion -- \textit{easy} and \textit{hard} -- and by object size -- \textit{small}, \textit{medium}, and \textit{large}. We report results for each subset following the HR-InsDet evaluation protocol.
The second benchmark, Robotools~\cite{robotools}, contains 20 object instances and 1581 test images from 24 indoor scenarios. Unlike HR-InsDet, Robotools prohibits using its 20 search targets for training, thereby evaluating generalization to novel objects. Accordingly, we train CF-GAP using only objects from HR-InsDet.
We report average precision (AP) at Intersection-over-Union (IoU) thresholds from 0.5 to 0.95 in steps of 0.05, as well as AP$_{50}$ at an IoU threshold of 0.5.

\subsubsection{Baselines.}

The strongest baselines, OTS-FM~\cite{hr_ins_det}, IDOW~\cite{shen2025solving}, and NIDS-Net~\cite{nidsnet}, employ pre-trained foundation models to process the entire scene, first detecting bounding boxes for all object-like regions (proposals). A feature extractor then generates embeddings for each proposal, which are matched to the search target's examples via Stable Matching~\cite{stable_matching}.
All methods use either SAM~\cite{sam} or GroundingDINO~\cite{grounding_dino} for proposals and DINOv2~\cite{dinov2} for feature extraction.
IDOW extends OTS-FM by fine-tuning DINOv2 on search targets from HR-InsDet. However, we exclude IDOW from our evaluations as its fine-tuned weights are unavailable, precluding integration with our CF-GAP front-end.
We do include NIDS-Net, which follows a similar but higher-performing approach: it trains a weight adapter for DINOv2 and additionally uses SAM to mask out backgrounds within each proposal.
The core baseline set comprises OTS-FM$_{\text{SAM}}$ and OTS-FM$_{\text{GroundingDINO}}$, depending on proposal detector, and NIDS-Net, which uses GroundingDINO for proposal detection.
We additionally consider two efficient SAM variants as OTS-FM backbones: MobileSAM~\cite{mobilesam}, a distilled version of SAM, and Segment This Thing (STT)~\cite{stt}, which uses a fovea-inspired tokenization that partitions the image into patches of increasing size with distance from a given location.
These are denoted OTS-FM$_{\text{MobileSAM}}$ and OTS-FM$_{\text{STT}}$.
Since STT requires a location input for the tokenization, it can only be evaluated in combination with CF-GAP and is thus excluded from the main pairwise comparisons between standalone baselines and their CF-GAP extensions. Unless stated otherwise, all baseline results are reproduced using the publicly available code.

\subsubsection{Setup.}
We integrate CF-GAP with each baseline as its downstream architecture.
For the \bMobileSAM~and \bSAM~baselines, CF-GAP changes the input structure: standalone baselines receive the full high-resolution scene together with a coarse grid of 2D point prompts to detect proposals at each grid location. By contrast, CF-GAP provides only a small RoI along with fine glimpse locations as 2D point prompts at the end of each fine glimpsing loop, improving both efficiency and focus.
As the \bGDINO~baseline does not support point-based prompting, the fine glimpse locations are used to filter out detected bounding boxes that do not contain them.

In HR-InsDet and Robotools, scenes are sized at $6144 \times 8192$ and $1920 \times 1080$ pixels, respectively, whereas baseline models require resizing below $2048 \times 2048$.
CF-GAP supports flexible input sizes; for faster experimentation, HR-InsDet scenes were resized to $4096 \times 5460$ while Robotools scenes were kept at original resolution.
Coarse glimpsing operates on scenes downscaled by a factor of~2; fine glimpsing operates at full resolution.
The log-polar sensor diameter is set to 4096 pixels, and the resulting glimpses are resized to $245 \times 245$.
Each scene undergoes $N_c$ coarse glimpses, each followed by $N_f$ fine glimpses, with $N_c{=}30$ and $N_f{=}3$ by default unless stated otherwise.
Further details are provided in Appendix B.

\section{Results}\label{sec:Results}

\subsection{Benchmarking results}
As shown in \cref{fig:MainResult,,fig:MainResultRobotools}, CF-GAP consistently improves all baseline models, demonstrating the effectiveness of the coarse-to-fine glimpsing.
In evaluations on HR-InsDet dataset, the biggest benefits are observed in scenes with small-sized objects and hard scenes, \ie scenes with high clutter and partial occlusions.
\bMobileSAM, the smallest baseline, benefits the most, achieving up to 20\% AP improvement on hard scenes.
With CF-GAP, it surpasses both \bSAM~and \bGDINO~on the difficult subsets while performing comparably on simpler ones.
On Robotools, CF-GAP also consistently improves AP of all baselines. Lightweight \bMobileSAM~paired with CF-GAP achieves AP$_\text{50}$ comparable even with the heavier baselines; its AP, however, stays below theirs, indicating less precise bounding boxes due to its weaker detector backbone.
We provide more extensive tabular comparisons in Appendix C.
\MainResult{!t} 
\MainResultRobotools{!t} 

\subsection{Analysis and Ablations}
\subsubsection{Naive high-resolution baseline.} 
To demonstrate the importance of CF-GAP, we compare it against a straightforward alternative for processing high-resolution images.
Baseline models cannot handle scenes at their original resolution because their ViT-based backbones decompose images into a fixed number of patches to limit the quadratic cost of self-attention.
A naive solution is to split each scene into smaller overlapping patches and let the baseline treat each patch as a separate RoI.
\cref{tab:PatchAblationTable} reports results for this approach, where the full-sized ($6144 \times 8192$) scenes are split into $1024 \times 1024$ patches with 50\% overlap, yielding 165 RoIs per scene -- over $5\times$ more than the 30 RoIs produced by CF-GAP.
OTS-FM models without fine-tuned matching do not benefit from such patching, as the larger number of candidate proposals across all RoIs leads to increased matching errors.
NIDS-Net, whose fine-tuned features better discriminate the search target, does benefit -- particularly for small objects -- but still falls behind its CF-GAP extension, especially in hard, cluttered scenes.
These results confirm that while exhaustive patching can in principle recover lost high-resolution detail, CF-GAP is both more effective -- achieving higher AP, and more efficient -- passing over $5\times$ fewer RoIs to the downstream architecture.
\PatchAblationTable{!t}

\subsubsection{Glimpse locations.}
In addition to providing targeted RoIs to downstream architectures, CF-GAP also provides fine glimpse locations that specify where objects have to be detected within each RoI. For SAM-based models, this improves the efficiency by replacing the dense grid of point prompts with only a few glimpse locations. 
For GroundingDINO-based models, the fine glimpse locations act as spatial filters that exclude candidate objects that do not overlap with them, substantially reducing the number of candidates and thereby improving matching effectiveness.
This is confirmed in \cref{tab:LocationForDA}, which reports higher performance when the downstream architecture receives both the RoIs and the glimpse locations, compared to receiving only the RoIs.

\LocationForDA{!h}

\subsubsection{Fine glimpsing.}
Given that CF-GAP consists of coarse and fine glimpsing processes, it is important to demonstrate the need for the latter. 
\cref{tab:FineGapAblationTable}~shows that fine glimpsing is most beneficial in hard scenes and scenes with small-sized objects. 
This is because the low resolution of the coarse search map can yield glimpse locations that are offset from the actual object, disrupting downstream architectures that are prompted to detect objects at specific locations. 
Fine glimpsing corrects for this spatial error (see \cref{fig:QualitativeGlimpses} for examples).
In simpler cases where objects are easy to find, fine glimpsing provides marginal benefit.
\FineGapAblationTable{!h}

\subsubsection{Log-polar glimpses.}
\LogPolarQualitative{!b}
\LogPolarVSCropPerformance{!t}
Another study shows the advantage of extracting log-polar glimpses. 
Compared to a naive approach of extracting a small cartesian crop, the log-polar representation eliminates the need to tune the crop size, which would otherwise be highly sensitive to the proximity of the initial coarse glimpse location to the object and to the object's size. For example, the crop size can be either too small, providing too little information, or too big, providing too much distraction (\cref{fig:LogPolarQualitative}, top row). 
By contrast, due to logarithmically diminishing resolution at the periphery, the log-polar glimpses consistently maintain the focus on nearby regions regardless of the area captured in the full-resolution scene. 
This is visually apparent in the bottom row of \cref{fig:LogPolarQualitative}, where log-polar glimpses remain visually similar regardless of their size (diameter), as opposed to crop-based glimpses in the top row. 
As a result, the fine search map extraction becomes more robust against the distracting peripheral information as indicated by the resulting fine glimpse locations in \cref{fig:LogPolarQualitative}. 
\cref{fig:LogPolarVSCropPerformance} confirms this quantitatively, showing more stable performance over a broader range of glimpse sizes for log-polar compared to cropped-based glimpses. 
In addition, in Appendix D, we empirically justify our architecture for processing log-polar glimpses (\cref{fig:FineSearchMap}B-C) by comparing it to a CNN-based model, showing that the latter is less effective.

\subsubsection{Factoring out the downstream architecture.}
Instance detection can be decomposed into two stages: 1) finding a candidate region likely to contain the search target, and 2) matching it against the search target's examples for recognition.
Since CF-GAP primarily improves the first stage, we measure how often the search target is found in the scene regardless of whether it is successfully recognized during matching.
We note that we cannot report more standard average recall metrics, as CF-GAP does not directly detect bounding boxes.
\MissErrorAnalysisTable{!b}
Focusing on the most challenging HR-InsDet subsets, \cref{table:MissErrorAnalysisTable} shows that CF-GAP locates small objects and objects in hard scenes more frequently than standalone baselines.
These results also reflect the upper-bound AP achievable under a perfect matching stage.
The last row of \cref{table:MissErrorAnalysisTable} further highlights the importance of fine glimpsing for precise localization.

\QualitativeGlimpses{!t}
\subsubsection{Qualitative inspection.}
We provide visualizations of coarse and fine glimpse locations for a couple of scenes in \cref{fig:QualitativeGlimpses}. Note that the glimpses are not spread out across entire scenes meaning that only a subset of the entire scene will be passed to downstream architectures. This, in turn, implies the reduction of irrelevant and potentially distracting information. 
The visualizations in zoom-in panels also illustrate how fine glimpsing corrects for the spatial imprecision of coarse glimpsing. More examples are provided in Appendix E. Failed cases are shown and discussed in Appendix F.

\subsection{Computational Cost}
\EfficiencyAnalysis{!t}
\GlimpseAnalysis{!b}

While CF-GAP significantly improves performance of the baselines, it incurs additional computational cost of repeatedly running the downstream architecture after each coarse glimpse.
To fully leverage the power of CF-GAP, one has to select the downstream architecture wisely.
In particular, it is costly to use a heavy, inefficient model designed and trained to handle complex scenery with numerous objects and various visual intricacies. In fact, this is unnecessary since CF-GAP provides targeted, high-quality information stripped of irrelevant details. Hence, a less powerful but more efficient downstream architecture can suffice.    
To illustrate this, we compare SAM with its two more efficient versions -- MobileSAM and STT -- as OTS-FM's backbones combined with CF-GAP. The comparison is made both in terms of performance and efficiency, with the latter being represented via end-to-end FLOPS for all models. 
As can be seen in \cref{fig:EfficiencyAnalysis}, CF-GAP makes the more efficient versions surpass the original \bSAM~model in terms of both efficiency and performance. 
Although, to the best of our knowledge, there are no efficient versions currently available for GroundingDINO, we expect to observe similar trends to those of SAM and its efficient alternatives. 
\FlopsReport{!b}

More generally, with each coarse glimpse invoking the downstream architecture, CF-GAP directly trades efficiency for performance. \cref{fig:GlimpseAnalysis} traces this trade-off along the number of coarse glimpses $N_c$.  
The comparison between the standalone baseline and its CF-GAP extension at the matched compute budget corresponds to the case of using a single coarse glimpse.  
In addition, we break down the compute cost in \cref{tab:flops_report}, showing that the total cost is dominated by $N_c$ invocations of the downstream architecture. We report the cost in FLOPS, since wall-clock runtime depends on implementation-specific optimizations beyond the scope of this work.  
We also note that two further costs, shared by all downstream architectures, are omitted from the table: encoding the search target examples and the candidate objects with DINOv2, both of which vary across datasets and scenes.

Lastly, ~\cref{fig:GlimpseAnalysis} shows the cumulative percentage of hitting the search target (in gray): in $\sim$50\% of scenes, CF-GAP finds search targets within the first 8-10 glimpses.
This hints that the computational cost can be reduced, given a more robust matching stage that could terminate the glimpsing process once the search target is recognized.

\section{Discussion}

Our results demonstrate that CF-GAP consistently boosts existing instance detectors, with the largest gains in the most challenging settings of small-sized objects and cluttered scenes.
Combining CF-GAP with instance detectors as downstream architectures exhibits a functional dichotomy of looking and seeing: CF-GAP \textit{looks} for task-relevant regions and directs the downstream architecture as a \textit{seeing} component to analyze them in high resolution. 
This division of labor, in turn, allows for employing lighter, distilled models such as \bMobileSAM~as a downstream architecture, achieving competitive performance compared to larger models. 
Moreover, one can also use large downstream architectures such as \bSTT~with advanced fovea-inspired tokenization techniques that allow to retain their expressivity while making them very efficient.  
Hence, the combination of our lightweight CF-GAP with such downstream architectures paves the way to efficient yet powerful task-driven models. 

\noindent\textbf{Limitations and future work.} In its current form, CF-GAP relies solely on texture-based guidance. However, the human visual system is known to leverage high-level semantics about objects, spatial layouts of scenes, and many other features when searching for task-relevant information. 
In addition, the inhibition-of-return mechanism only suppresses previously visited locations and their immediate neighborhoods, rather than entire task-irrelevant regions.
This can cause the glimpsing process to repeatedly revisit the same distractor object.
Another limitation is that CF-GAP passes a fixed-size crop as RoI to the downstream architecture, requiring the crop to be conservatively large to accommodate objects of varying sizes.
An adaptive mechanism that adjusts the crop based on the content of the task-relevant region would improve both efficiency and precision.
Finally, invoking the downstream architecture after every coarse glimpse incurs a computational cost that grows linearly with the number of coarse glimpses. A more robust matching stage that halts the glimpsing process once the search target is confidently recognized would alleviate this cost overhead.
Addressing these limitations and extending CF-GAP to other task definitions, such as text-based queries, are promising directions for future work.

%
%
\bibliographystyle{splncs04}
\nocite{cut_paste_learn, fastercnn, retinanet, centernet, fcos, cpldino, os2d, dtoid, oln, esteves2017polar, vaswani2017attention, dosovitskiy2021an, deformed_convs}
\bibliography{main}

\newcommand{\FullResultsTableHRDataset}[1]{
\begin{table*}[#1]\label{tab:fullresults}
\caption{Performance on HR-InsDet for all our models and models evaluated in prior work.
Results for models that were subsequently evaluated with CF-GAP are reproduced using publicly available code.
Note that OTS-FM$_\text{STT}$ can be evaluated only in combination with CF-GAP as its STT backbone requires glimpse locations as inputs.
}
\centering
\resizebox{\textwidth}{!}{
\begin{tabular}{l|l|cccccc|c} 
\multirow{2}{*}{Model}                                 & \multirow{2}{*}{Venue and Year} & \multicolumn{6}{c|}{AP}               & \multirow{2}{*}{AP$_{50}$}  \\
                                                                 &                                      & avg   & hard  & easy  & small & medium & large &                                     \\ 
\hline
CPL$_\text{FasterRCNN}$ \cite{cut_paste_learn, fastercnn}                                  & NeurIPS 2015                         & 19.5 & 10.3 & 23.8 & 5.0  & 22.2  & 38.0 & 29.2 \\
CPL$_\text{RetinaNet}$ \cite{cut_paste_learn, retinanet}                                     & ICCV 2017                            & 22.2 & 14.9 & 26.5 & 5.5  & 25.8  & 42.7 & 31.2 \\
CPL$_\text{CenterNet}$ \cite{cut_paste_learn, centernet}                                     & CVPR 2019                            & 21.1 & 11.9 & 25.7 & 5.9  & 24.2  & 40.4 & 32.7 \\
CPL$_\text{FCOS}$ \cite{cut_paste_learn, fcos}                                          & ICCV 2019                            & 22.4 & 13.2 & 28.7 & 6.2  & 26.5  & 38.1 & 32.8 \\
CPL$_\text{DINO}$ \cite{cut_paste_learn, cpldino}                                        & ICLR 2023                            & 28.0 & 17.9 & 32.7 & 11.5 & 31.5  & 48.4 & 39.6 \\
\hline
OTS-FM$_\text{MobileSAM}$ \cite{hr_ins_det, mobilesam}                                               & NeurIPS 2023                         & 37.0 & 22.0 & 43.1 & 12.4 & 42.4  & 63.2 & 46.1 \\
OTS-FM$_\text{SAM}$ \cite{hr_ins_det, sam}                                               & NeurIPS 2023                         & 41.6 & 28.0 & 47.6 & 14.6 & 45.8  & 69.1 & 49.1 \\
OTS-FM$_\text{GroundingDINO}$ \cite{shen2025solving, grounding_dino}                          & CVPR 2025                            & 51.7 & 37.2 & 58.7 & 28.8 & 58.6  & 69.2 & 62.5 \\
IDOW$_\text{SAM}$  \cite{shen2025solving, sam}                                              & CVPR 2025                            & 48.8 & 32.1 & 56.5 & 20.8 & 55.3  & 73.4 & 49.1 \\
IDOW$_\text{GroundingDINO}$ \cite{shen2025solving, grounding_dino}                                   & CVPR 2025                            & 57.0 & 40.7 & 64.4 & 35.3 & 63.0  & 73.6 & 69.3 \\
NIDS-Net \cite{nidsnet}                                   & IROS 2025                            & 63.8 & 39.9 & 74.6 & 32.4 & 72.7  & 86.1 & 69.8 \\
\hline
\hline

CF-GAP + OTS-FM$_\text{SAM}$                                      &           & 56.5     & 47.1     & 60.7     & 32.4     & 62.6      & 75.6 & 61.0   \\
CF-GAP + OTS-FM$_\text{GroundingDINO}$ &                                      & 58.8     & 50.2     & 62.7     & 39.6     & 63.5      & 74.8 & 63.0    \\
CF-GAP + OTS-FM$_\text{MobileSAM}$     &                                      & 51.8     & 41.7     & 56.4     & 29.3     & 57.9      & 68.6 & 58.1    \\
CF-GAP + OTS-FM$_\text{STT}$           &                                      & 52.6     & 42.7     & 57.0     & 30.2     & 58.8      & 68.8 & 60.5    \\
CF-GAP + NIDS-Net &                                      & \textbf{73.3}     & ~\textbf{63.2}    & \textbf{78.2}     & \textbf{52.1}     & \textbf{79.7}      & \textbf{86.1} & ~\textbf{78.4}   \\
\end{tabular}
}
\end{table*}
}

\newcommand{\FullResultsTableRobotools}[1]{
    \begin{table*}[#1]
    \caption{Performance on Robotools for all our models and models evaluated in prior work. 
    }
    \label{tab:robotools}
    \centering
    \resizebox{0.7\columnwidth}{!}{%
    \begin{tabular}{l|l|cc}
    Model       & Venue and Year                & AP   & AP$_{50}$  \\ 
    \hline
    OS2D  \cite{os2d}      &  ECCV 2020  & 2.9 & 6.5    \\
    DTOID \cite{dtoid}       &  WACV 2021  & 3.6 & 9.0    \\
    OLN$_{\text{Corr}}$ \cite{oln} &  RA-L 2022  & 14.4 & 18.1    \\
    VoxDet \cite{robotools} &  NeurIPS 2023  & 18.7 & 23.6    \\
    \bMobileSAM   \cite{hr_ins_det, mobilesam}    &    NeurIPS 2023       & 35.5 & 46.9    \\
    \bSAM   \cite{hr_ins_det, sam}    &    NeurIPS 2023       & 46.5 & 55.9    \\
    \bGDINO   \cite{shen2025solving, grounding_dino} &   CVPR 2025     & 56.7 & 64.8    \\
    IDOW$_{\text{SAM}}$   \cite{shen2025solving, sam}    &    CVPR 2025         & 51.9 & 63.8    \\
    IDOW$_{\text{GroundingDINO}}$ \cite{shen2025solving, grounding_dino}  &   CVPR 2025    & 59.0 & 67.8   \\
    NIDS-Net \cite{nidsnet}  &   IROS 2025    & 64.9 & \textbf{79.4}   \\
    \hline \hline
    CF-GAP + \bSAM     &      & 55.2 & 68.1    \\ 
    CF-GAP + \bGDINO &        & 62.1 & 70.8    \\ 
    CF-GAP + \bMobileSAM  &    & 49.4 & 63.5    \\ 
    CF-GAP +  OTS-FM$_\text{STT}$     &   & 46.5 & 59.9    \\ 
    CF-GAP +  NIDS-Net     &   & \textbf{70.3} & \textbf{79.4}    \\ 
    \end{tabular}
    }
    \end{table*}
}

\newcommand{\CNNJustificationTable}[1]{
    \begin{table}[#1]
    \centering
    \caption{Performance comparison when using CNN-based model for scene and search target encoders. Results are shown for CF-GAP+NIDS-Net evaluated on the two most challenging HR-InsDet subsets.    }
    \label{table:CNNJustificationTable}
    \begin{tabular}{c|cc}
    \hline
    \multirow{2}{*}{Scene \& search target encoders}   & \multicolumn{2}{c}{AP}  \\
                                                & small & hard            \\ 
    \hline
    \multicolumn{1}{l|}{CNN}                    & 47.5  & 58.6            \\
    \multicolumn{1}{l|}{Cross- \& self-attention blocks} & \textbf{52.1}  & \textbf{63.2}            \\
    \hline
    \end{tabular}
    \end{table}
}

\clearpage

\appendix

\section{Model Details}\label{appendix:model_details}
The code can be accessed via this \href{https://ibm.biz/coarse-to-fine-gap}{link}.
\subsection{Log-polar sensor} 
\begin{figure}[h]
    \centering
    \includegraphics[width=0.7\linewidth]{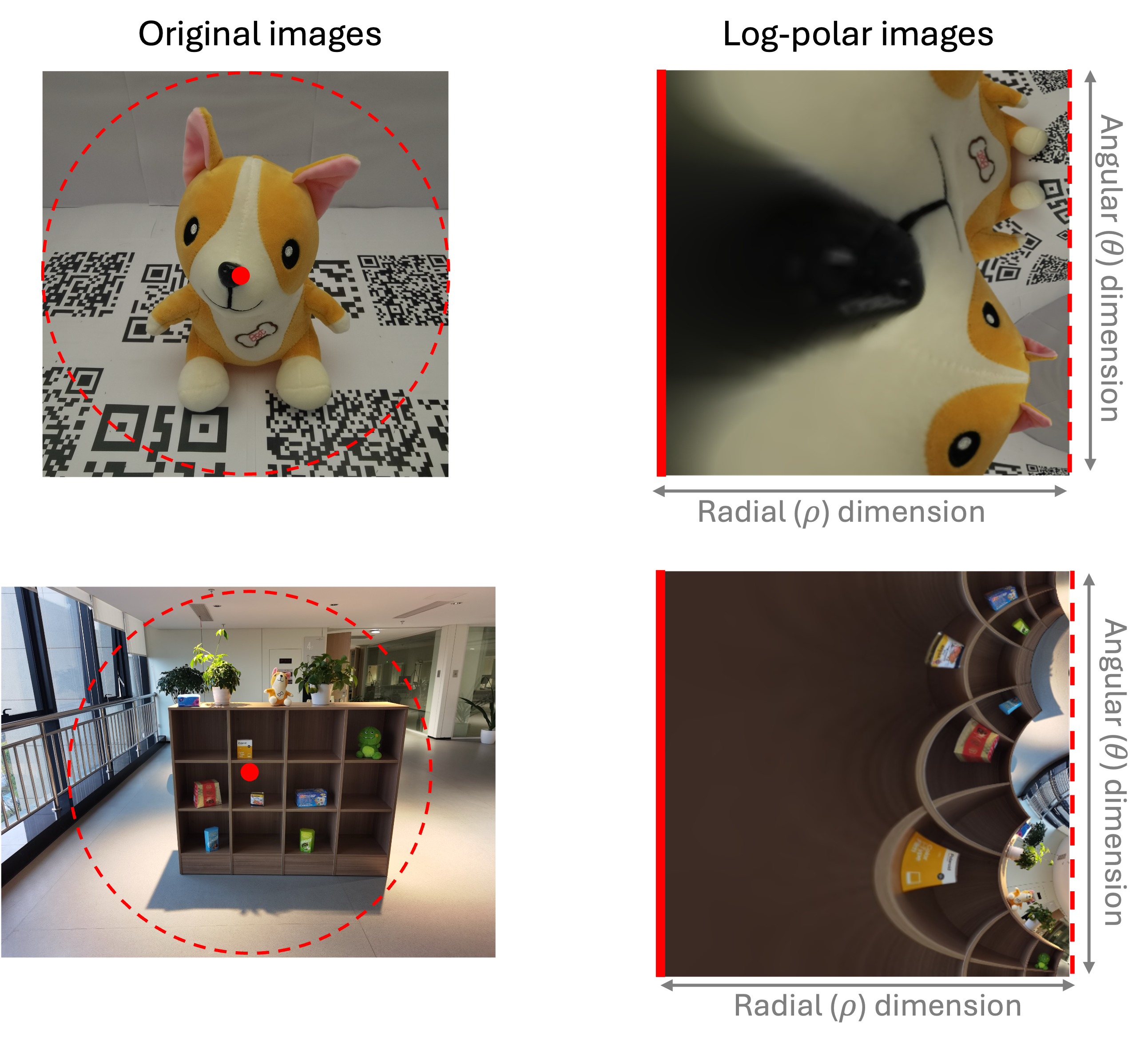}
    \caption{
        Log-polar images produced around the red dots in the original images, with the dashed contours corresponding to the border of the area from which the pixels are sampled. Note that the red dots and dashed contours in the original images are mapped to the red solid and dashed lines, respectively, in the log-polar images.
    }
    \label{fig:logpolar}
\end{figure}

The log-polar sensor samples pixels based on the log-polar layout around the glimpse location, oversampling regions that are closer to the glimpse location and undersampling ones that are farther.
More specifically, given an image $\boldsymbol{I}$ of size $H \times W$ and a glimpse location $(x, y)$, the sensor samples pixels from the image $\boldsymbol{I}$ according to the log-polar coordinate transform as in \cite{esteves2017polar}:
\begin{align}\label{eq:logpolar}
x_s = x + e^{\log(\rho)x_t/W} \cos(\frac{2\pi y_t}{H}) \\
y_s = y + e^{\log(\rho)x_t/W} \sin(\frac{2\pi y_t}{H})
\end{align}
where $(x_s, y_s)$ denote the sampled points from $(x_t, y_t)$ coordinates in the image $\boldsymbol{I}$, and $\rho$ is a hyper-parameter that defines the radius of the region around the glimpse location in $\boldsymbol{I}$ from which the pixels are to be sampled. 
Compared to the cartesian grid-based layout, the log-polar design offers a better resolution vs. field of view balance. In particular, it magnifies regions near glimpse locations, see \cref{fig:logpolar} for an illustration.

\subsection{Fine search map}\label{subsec:fine_search_map}

\subsubsection{Input preparation.}
The scene and search target encoders for fine search map generation, shown in~\cref{fig:FineSearchMap}B-C, receive a scene glimpse $\boldsymbol{G}_s \in \mathbb{R}^{H_\theta \times W_\rho \times 3}$ and a sequence of $M$ search target glimpses $\{\boldsymbol{G}_m\}_{m=1}^{M}$, $\boldsymbol{G}_m \in \mathbb{R}^{H_\theta \times W_\rho \times 3}$, respectively. Each glimpse is an RGB log-polar image of size $H_\theta \times W_\rho$, where $H_\theta$ is the angular dimension and $W_\rho$ is the radial dimension (see \cref{fig:logpolar}).
The search target glimpses are extracted from an image of the search target by applying the log-polar sensor at the center of the object as well as at its topmost and bottommost locations, yielding $M=3$ glimpses. These locations are determined from the binary segmentation mask provided for each search target image. During training, a fourth glimpse ($M{=}4$) is additionally sampled at a random target location to define the localization target, as detailed in \cref{subsec:training_details}.

All glimpses are partitioned into non-overlapping patches of size $P{\times}P$, each of which is projected into a $D$-dimensional feature space.
This produces the scene feature grid $\boldsymbol{F}_s \in \mathbb{R}^{H'_\theta \times W'_\rho \times D}$ and the search target feature grids $\{\boldsymbol{F}_m\}_{m=1}^{M}$, $\boldsymbol{F}_m \in \mathbb{R}^{H'_\theta \times W'_\rho \times D}$, where the new size $H'_\theta\times W'_\rho$ accounts for patching. Note that in \cref{sec:Model} the subscripts $\cdot_\theta$ and $\cdot_\rho$ were omitted for the sake of simplicity. In the technical description below, we restate Eqs.~\ref{eq:scene_compress_m}-\ref{eq:target_cross_m} for the reader's convenience.

\subsubsection{Scene encoder.}
The scene encoder (\cref{fig:FineSearchMap}B) receives the scene feature grid $\boldsymbol{F}_s$ as input and outputs a glimpse feature map $\boldsymbol{F}^*_s \in \mathbb{R}^{H'_\theta \times W'_\rho \times D}$ through three stages: compression, processing, and decoding.

\paragraph{Compression.}
The feature grid $\boldsymbol{F}_s$ is first flattened into a sequence of $H'_\theta \times W'_\rho$ tokens and compressed into a compact set of $N$ latent embeddings via cross-attention:
\begin{equation}\label{eq:scene_compress}
    \tilde{\boldsymbol{h}}_s = \mathcal{C}\!\left(\mathsf{Q} = \boldsymbol{e}_q,\; \mathsf{KV} = \boldsymbol{F}_s\right),
\end{equation}
where $\boldsymbol{e}_q \in \mathbb{R}^{N \times D}$ is a set of learnable embeddings. Since $N \ll H'_\theta \times W'_\rho$, this step serves as a bottleneck that alleviates the quadratic cost of the subsequent self-attention.

\paragraph{Processing.}
The latent embeddings $\tilde{\boldsymbol{h}}_s \in \mathbb{R}^{N \times D}$ are subsequently processed through self-attention:
\begin{equation}\label{eq:scene_process}
    \boldsymbol{h}_s = \mathcal{S}\!\left(\mathsf{QKV} = \tilde{\boldsymbol{h}}_s\right),
\end{equation}
where $\boldsymbol{h}_s \in \mathbb{R}^{N \times D}$. This stage enables the integration of fine local details encoded near the glimpse center with the broader peripheral context.

\paragraph{Decoding.}
The processed latent embeddings $\boldsymbol{h}_s$ are decoded back to the original spatial resolution using another cross-attention and scene feature grid $\boldsymbol{F}_s$ as queries:
\begin{equation}\label{eq:scene_decode}
    \boldsymbol{F}^*_s = \mathcal{C}_{\mathrm{dec}}\!\left(\mathsf{Q} = \boldsymbol{F}_s,\; \mathsf{K} = \boldsymbol{h}_s,\; \mathsf{V} = \boldsymbol{e}_o\right),
\end{equation}
where $\boldsymbol{e}_o \in \mathbb{R}^{N \times D}$ is a distinct set of learnable embeddings that defines the output space of the final glimpse feature map $\boldsymbol{F}^*_s$. 

Throughout, $\mathcal{C}(\cdot)$ and $\mathcal{S}(\cdot)$ denote cross-attention and self-attention blocks, respectively, each implemented as a single-layer transformer~\cite{vaswani2017attention}. These blocks are shared with the search target decoder described below. The decoding cross-attention $\mathcal{C}_{\mathrm{dec}}(\cdot)$ uses a separate set of parameters.

\subsubsection{Search target encoder.}
The search target encoder (\cref{fig:FineSearchMap}C) iteratively processes the sequence of search target feature grids $\{\boldsymbol{F}_m\}_{m=1}^{M}$, $\boldsymbol{F}_m \in \mathbb{R}^{H'_\theta \times W'_\rho \times D}$, through cross-attention and self-attention blocks that are shared with the scene encoder, \cref{eq:scene_compress,eq:scene_process}.
At each iteration $m$, the cross-attention block receives the flattened feature grid $\boldsymbol{F}_m$ as keys and values, and the output of the previous iteration as queries:
\begin{align}
    \tilde{\boldsymbol{h}}_T^{(m)} &= \mathcal{C}\!\left(\mathsf{Q} = \boldsymbol{h}_T^{(m-1)},\; \mathsf{KV} = \boldsymbol{F}_{m}\right), \label{eq:target_cross} \\
    \boldsymbol{h}_T^{(m)} &= \mathcal{S}\!\left(\mathsf{QKV} = \tilde{\boldsymbol{h}}_T^{(m)}\right), \label{eq:target_self}
\end{align}
where $\tilde{\boldsymbol{h}}_T^{(m)}, \boldsymbol{h}_T^{(m)} \in \mathbb{R}^{N \times D}$.
At the first iteration ($m{=}1$), the query is initialized with the same learnable embeddings used by the scene encoder, \ie $\boldsymbol{h}_T^{(0)} = \boldsymbol{e}_q$.
Unlike the scene encoder, the search target encoder omits the decoding stage; instead, the output of the final iteration $\boldsymbol{h}_T = \boldsymbol{h}_T^{(M)} \in \mathbb{R}^{N \times D}$ is used directly as the set of target features for feature comparison.

\subsubsection{Feature comparison.}
The target features $\boldsymbol{h}_T \in \mathbb{R}^{N \times D}$ are spatially correlated with the glimpse feature map $\boldsymbol{F}^*_s \in \mathbb{R}^{H'_\theta \times W'_\rho \times D}$ to produce the fine search map. Each of the $N$ target embeddings $\boldsymbol{h}_T^{(n)} \in \mathbb{R}^{D}$, $n = 1, \dots, N$, is treated as a $1{\times}1$ convolution kernel and convolved over $\boldsymbol{F}^*_s$, yielding $N$ correlation maps:
\begin{equation}\label{eq:correlation}
    \boldsymbol{C}_n(i,j) = \boldsymbol{h}_T^{(n)} \cdot \boldsymbol{F}^*_s(i,j), \quad \boldsymbol{C}_n \in \mathbb{R}^{H'_\theta \times W'_\rho},
\end{equation}
where $\cdot$ denotes the dot product. These correlation maps are averaged into a single fine search map $\boldsymbol{S}_{\text{fine}} = \frac{1}{N}\sum_{n=1}^{N} \boldsymbol{C}_n$.

To improve robustness to viewpoint variation, search target glimpses are extracted from four images of the search target, each depicting the object from a different viewpoint. The above procedure is applied independently to each viewpoint example so that the final fine search map is averaged across maps produced for each example.

\subsubsection{Positional Embeddings.}
While omitted in the description above, we add positional embeddings to the feature grids $\{\boldsymbol{F}_m\}_{m=1}^{M}$ and $\boldsymbol{F}_s$.  
We adapt the standard sinusoidal positional embeddings (SPEs) from \cite{vaswani2017attention, dosovitskiy2021an} to the log-polar grid. 
Let $\boldsymbol{F}\in\mathbb{R}^{H'_\theta\times W'_\rho \times D}$ be a log-polar feature grid of size $H'_{\theta} \times W'_{\rho}$, where $H'_{\theta}$ is the number of patches along the angular dimension and $W'_{\rho}$ is the number of patches along the radial dimension. 
For a patch at grid position $(p, q)$, where $p \in \{0, \dots, H'_{\theta}-1\}$ and $q \in \{0, \dots, W'_{\rho}-1\}$, we compute its $D$-dimensional positional embedding $PE_{(p,q)}$ by decomposing it into two independent 1D embeddings.
Specifically, we partition the embedding dimension $D$ into $D = D_{\theta} + D_{\rho}$. The final embedding is the concatenation of the $\theta$-dimension embedding and the $\rho$-dimension embedding:
\begin{equation}
    PE_{(p,q)} = [PE_{\theta}(p) \oplus PE_{\rho}(q)]
\end{equation}
where $\oplus$ denotes vector concatenation.
The $\theta$-dimension of the log-polar grid is cyclic, meaning that the position $p=0$ is adjacent to $p=H'_{\theta}-1$. Using the standard sinusoidal formula would create an artificial ``seam'', incorrectly signaling a large distance between these adjacent patches.
To resolve this issue, we define $PE_{\theta}$ with cyclic frequencies. Specifically, at each position $p$ and frequency index $k \in \{1, \dots, D_{\theta}/2\}$, the components of $PE_{\theta}$ are defined as:
\begin{align}
PE_{\theta}(p, 2(k-1)) &= \sin\left(\frac{p \cdot 2\pi k}{H'_{\theta}}\right) \\
PE_{\theta}(p, 2(k-1)+1) &= \cos\left(\frac{p \cdot 2\pi k}{H'_{\theta}}\right)
\end{align}
This formulation ensures that $PE_{\theta}(p) = PE_{\theta}(p + H'_{\theta})$, providing a continuous and cyclic representation of the angular position in the log-polar grid. 
$PE_{\rho}$ is defined by the standard 1D SPE: for a position $q$ and embedding dimension index $k' \in \{0, \dots, D_{\rho}/2 - 1\}$, the components are:
\begin{align}
PE_{\rho}(q, 2k') &= \sin(q / 10000^{2k'/D_{\rho}}) \\
PE_{\rho}(q, 2k'+1) &= \cos(q / 10000^{2k'/D_{\rho}})
\end{align}
The resulting positional embeddings are added to the log-polar feature grid $\boldsymbol{F}\in\mathbb{R}^{H'_\theta\times W'_\rho \times D}$.

\subsection{Location selection}
Coarse and fine glimpsing employ different mechanisms to select the next glimpse location based on their respective search maps. 

\subsubsection{Coarse glimpsing.}
At each coarse glimpse iteration $i=\{1, \dots, N_c\}$, a glimpse location $\boldsymbol{l}_i\in\mathbb{R}^2$ is selected using winner-takes-all (WTA), \ie picking the location in the coarse search map $S_c\in\mathbb{R}^{H_c \times W_c}$ with the highest value:
\begin{equation}
    \boldsymbol{l}_i = \text{WTA}(S_c^{(i)}) \overset{\text{def}}{=} \arg \max_{pq} (S^{(i)}_{c,~pq})
\end{equation}
where $S_c^{(i)}$ is the state of the coarse search map at iteration $i$.
The WTA operation is followed by inhibition-of-return (IoR), which applies a mask $M(\boldsymbol{l}_i) \in \mathbb{R}^{H_c \times W_c}$ around location $\boldsymbol{l}_i$ to prevent it from repetitive selection: 
\begin{equation}
    S_c^{(i+1)} = S_c^{(i)} \odot M(\boldsymbol{l}_i)
\end{equation}
where $\odot$ is the element-wise product and the value at each mask location $p,q$ is given by the inverted exponential radial kernel $1-e^{-\epsilon \| (p,q) - \boldsymbol{l}_i \|_2}$ with $\epsilon$ being a hyperparameter and $(p,q) \in \mathbb{R}^2$ a vector of location $(p,q)$.

\subsubsection{Fine glimpsing.}
At each fine glimpse iteration $j=\{1, \dots, N_f\}$, the fine glimpse location $\boldsymbol{l}_j^f=(c_\theta, c_\rho)$ is computed as a centroid of the fine search map $S_f \in \mathbb{R}^{H_\theta\times W_\rho}$, where $H_\theta$ and $W_\rho$ are the angular and radial dimensions of the log-polar image, respectively.
The fine search map is first normalized using spatial softmax so that the normalized values $\hat{S}_{f, ~pq}$ at each location $(p,q)$ become:
\begin{equation}
    \hat{S}_{f, ~pq} = \frac{\exp(S_{f, ~pq})}{\sum_{p'=1}^{H_\theta} \sum_{q'=1}^{W_\rho} \exp(S_{f, ~p'q'})}
\end{equation}
The radial index of the centroid $c_{\rho}$ is then defined as:
\begin{equation}
    c_{\rho} = \sum_{p=1}^{H_\theta} \sum_{q=1}^{W_\rho} q \cdot \hat{S}_{f, ~pq}
\end{equation}
To account for the periodicity of the angular dimension $H_\theta$, we define the angular position $\alpha_p$ for row $p$ of the log-polar image as $\alpha_p = 2\pi p / H_\theta$ and compute the weighted sum of sine and cosine components:
\begin{equation}
    \mathbf{v} = \sum_{p=1}^{H_\theta} \sum_{q=1}^{W_\rho} \hat{S}_{f, ~pq} 
    \begin{bmatrix} 
    \cos(\alpha_p) \\ 
    \sin(\alpha_p) 
    \end{bmatrix}
\end{equation}
The resulting angular centroid $c_{\theta}$ is calculated as:
\begin{equation}
    c_{\theta} = \frac{H_\theta}{2\pi} \left( \operatorname{atan2}(\mathbf{v}_y, \mathbf{v}_x) \pmod{2\pi} \right)
\end{equation}

\subsection{Training details}\label{subsec:training_details}
The model for coarse search map generation, MobileNetV3, is used with its pre-trained weights without any further fine-tuning.
The model for the fine search map generation is trained from scratch using a dataset synthesized with a cut-paste-learn (CPL) framework \cite{cut_paste_learn}. 
The synthetic scenes were generated by pasting original images of the search targets onto random backgrounds.
The training objective was to predict a search map that highlights a specific location of the search target within the synthetic scene. 

Given a synthetic scene $\boldsymbol{I}\in\mathbb{R}^{H \times W \times 3}$, we generate a scene glimpse $\boldsymbol{G}_s \in \mathbb{R}^{H_\theta\times W_\rho \times 3}$ by applying the log-polar sensor at a randomly selected location $\boldsymbol{l}_s \in\mathbb{R}^2$. Note that $H_\theta$ and $W_\rho$ denote the angular and radial dimensions of the resulting log-polar image, respectively.

Each training sample consists of a scene glimpse $\boldsymbol{G}_s$ and a sequence of $M=4$ search target glimpses $\{\boldsymbol{G}_m\}_{m=1}^{M}$, $\boldsymbol{G}_m \in \mathbb{R}^{H_\theta\times W_\rho \times 3}$. 
The first three glimpses ($m=1,2,3$) are generated by applying the log-polar sensor at the central, top, and bottom locations of the original search target image $\boldsymbol{I}_T$ to provide textural information. The fourth glimpse ($m=4$) is generated by applying the log-polar sensor at a randomly picked location $\boldsymbol{l}_t \in\mathbb{R}^2$ on the search target. This fourth glimpse specifies the exact part of the target to be localized within the scene glimpse $\boldsymbol{G}_s$.

The ground truth label is a spatial map $Y \in \mathbb{R}^{H_\theta\times W_\rho}$. This map consists of zeros except for a single element set to one at location $\boldsymbol{l}^*$. The coordinate $\boldsymbol{l}^*$ is obtained by mapping the randomly picked target location $\boldsymbol{l}_t$ (from $\boldsymbol{I}_T$) into the coordinate system of the scene glimpse $\boldsymbol{G}_s$. 
The model is trained using cross-entropy loss between the predicted flattened fine search map and the flattened one-ground truth map $Y$, treating each spatial location as a distinct class. Training is performed using objects and random backgrounds from the HR-InsDet dataset \cite{hr_ins_det}.

\section{Experimental details}\label{appendix:experimental_details}
In HR-InsDet and Robotools, scenes are sized at $6144 \times 8192$ and $1920 \times 1080$ pixels. While CF-GAP supports flexible image sizes, for faster experimentation, HR-InsDet scenes were resized to $4096 \times 5460$. The Robotools scenes were kept in original resolution. 
The coarse glimpsing operates on scenes downscaled by a factor of 2, while the fine glimpsing operates on full-resolution scenes. 
The size of the log-polar images is $H_\theta \times W_\rho=245 \times 245$, and the radius $\rho$ of the log-polar sensor is set to half of the smallest dimension of the scenes, \ie to 2048 and 540 for HR-InsDet and Robotools, respectively. 
Each scene undergoes $N_c=30$ coarse glimpses, with $N_f=3$ fine glimpses per coarse one. The size of the crop to be passed to the downstream architecture at the end each fine glimpsing loop is set $800 \times 800$ and $360 \times 360$ for HR-InsDet and Robotools, respectively.
The cross- and self-attention components of the fine search map are implemented as simple single-layer transformer networks with 4 heads and feature dimensionality set to $D=96$. The patching size is $P\times P = 5\times 5$, the number of learnable embeddings is $N=128$.

\FloatBarrier
\section{Additional Results}\label{appendix:additional_results}
\FullResultsTableHRDataset{!h}
\FullResultsTableRobotools{!h}

\begin{table}
    \caption{Sensitivity analysis of different hyperparameters. Results are shown for CF-GAP+NIDS-Net evaluated on the hard subset of HR-InsDet.}
    \label{tab:hp_sensitivity}
    \centering
    \resizebox{0.83\columnwidth}{!}{%
    \begin{tabular}{lcc}
    \toprule
    Hyperparameter & \multicolumn{1}{c}{Values} & \multicolumn{1}{c}{AP} \\
    \midrule
    IoR kernel $\epsilon$ & \{0.1 $\mid$ 0.5 $\mid$ 1 $\mid$ 10\} & \{63.0 $\mid$ \textbf{63.2} $\mid$ \textbf{63.2} $\mid$ 59.9\} \\
    \# fine glimpses $N_f$ & \{0 $\mid$ 1 $\mid$ 2 $\mid$ 3\} & \{56.2 $\mid$ 61.6 $\mid$ 62.7 $\mid$ \textbf{63.2}\} \\
    Log-polar radius $\rho$ & \{256 $\mid$ 512 $\mid$ 1024 $\mid$ 2048\} & \{60.9 $\mid$ 61.8 $\mid$ 61.8 $\mid$ \textbf{63.2}\} \\
    RoI size & \{$400^2$ $\mid$ $800^2$ $\mid$ $1200^2$ $\mid$ $1600^2$\} & \{57.4 $\mid$ \textbf{63.2} $\mid$ 60.4 $\mid$ 58.5\} \\
    \bottomrule
    \end{tabular}%
    }
\end{table}
\FloatBarrier
%

\section{Justifying the architecture for log-polar processing}\label{appendix:cnn_log_polar}
Generating high-quality fine search maps requires an architecture for scene and search target encoders that is suitable to process log-polar glimpses.  
As argued in \cref{sec:Model}, using CNN-based models for this purpose may not be the best solution due to non-uniform resolution within a log-polar image.  
We experimented with different types of CNN architectures and found particularly beneficial to use deformable convolutions \cite{deformed_convs} that can adjust the kernels depending on the spatial position of the receptive field within the log-polar image. 
However, even this best-performing CNN configuration lags behind the attention-based architecture shown in \cref{fig:FineSearchMap}B-C, as reported in \cref{table:CNNJustificationTable}.
We attribute this gap to the fact that the attention-based model allows for integrating fine local details near the glimpse location with the broader context of the more distant periphery.


\CNNJustificationTable{!h}


\section{Qualitative examples}\label{appendix:qualitative_examples}

\begin{figure}[!h]
    \centering
    \includegraphics[width=0.87\linewidth]{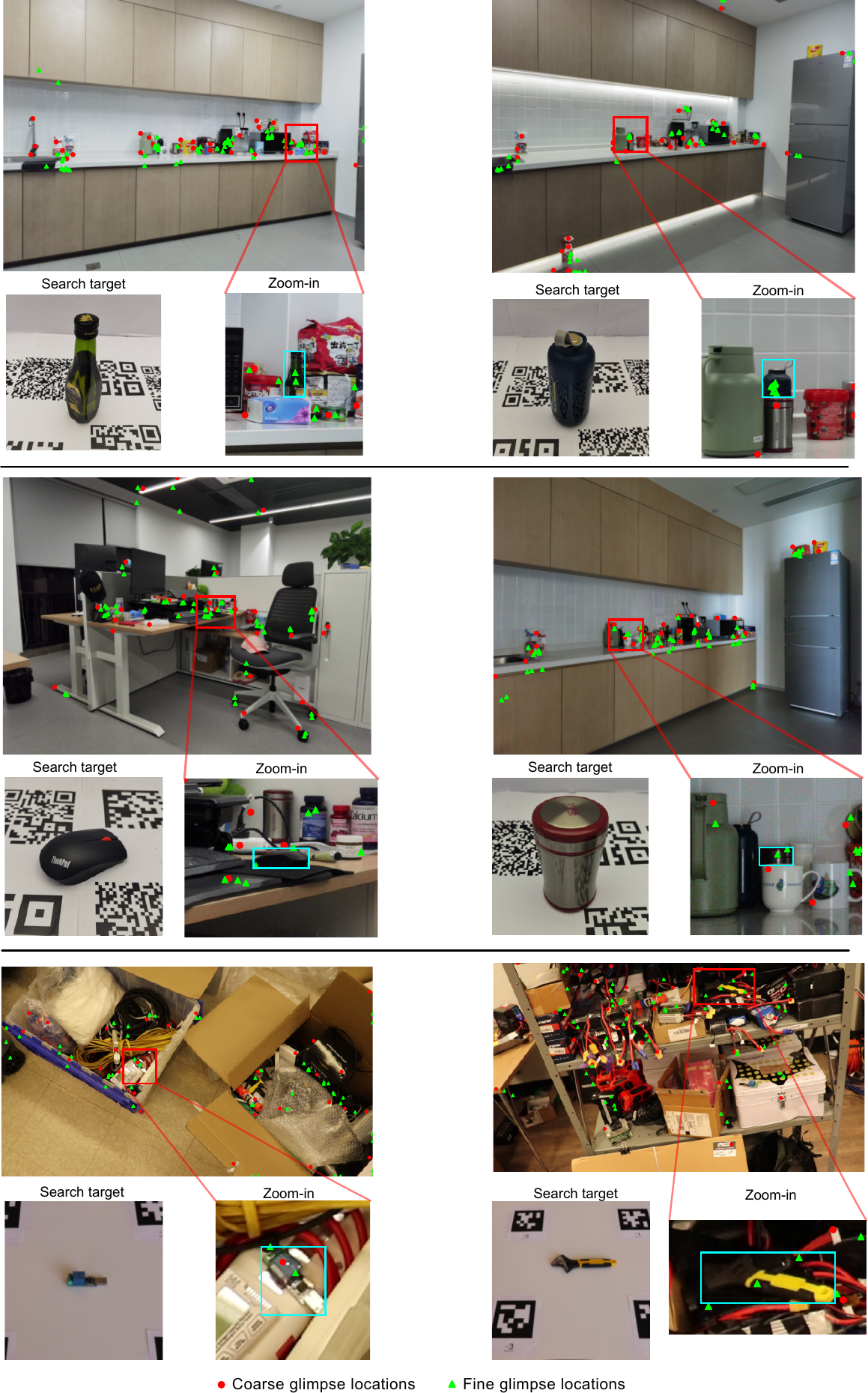}
    \caption{
        Coarse and fine glimpse locations for two scenes where baseline models failed to detect the search target, while our approach succeeded. Top four examples are from HR-InsDet, and two examples at the bottom are from Robotools.
    }
    \label{fig:QualitativeGlimpsesExt}
\end{figure}

\FloatBarrier
\section{Failed cases}\label{appendix:failed_cases}

\begin{figure}[!h]
    \centering
    \includegraphics[width=0.95\linewidth]{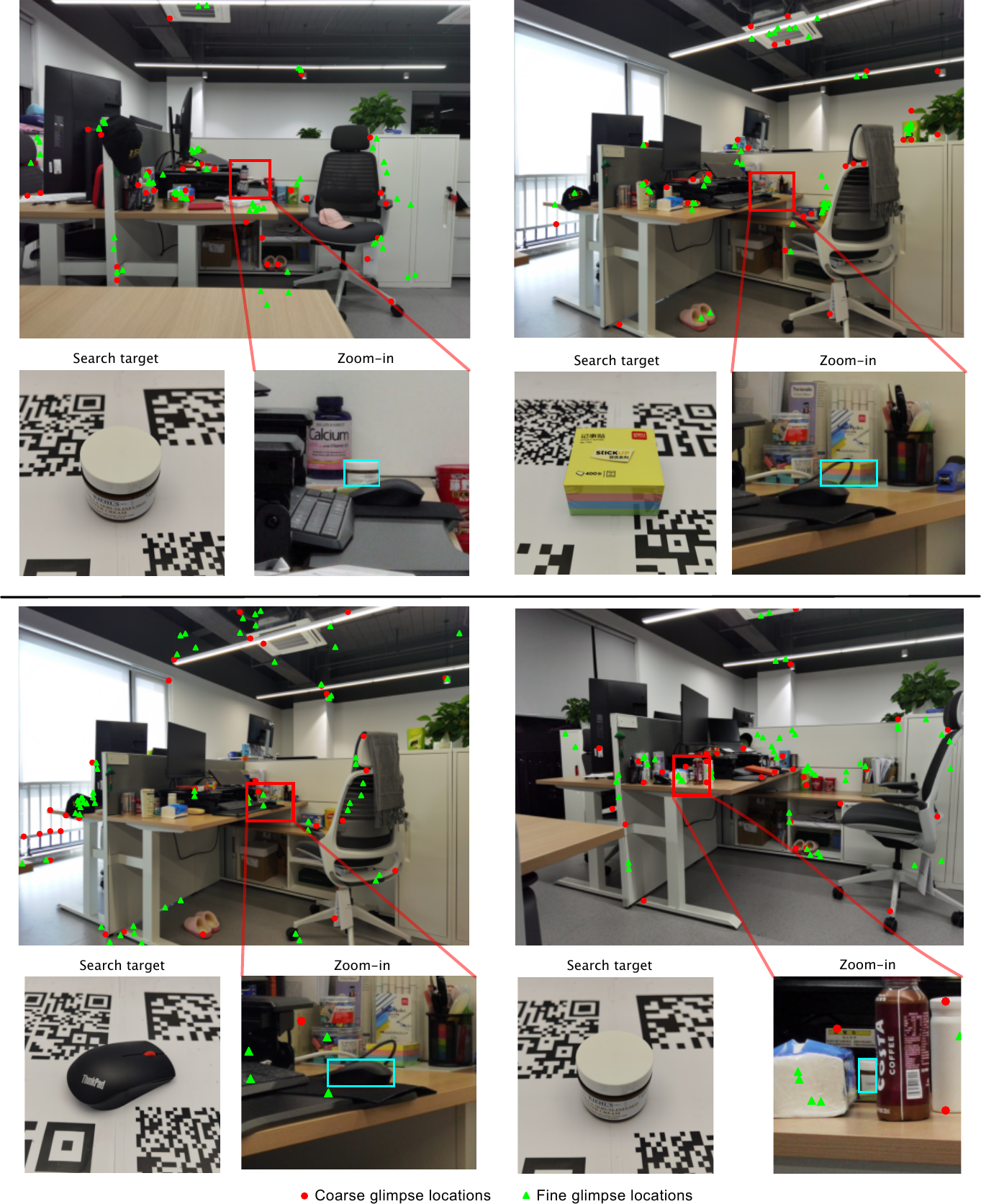}
    \caption{
        Failed cases where CF-GAP could not find the search target, \ie no glimpse location landed on the search target's surface.
    }
    \label{fig:failed_cases}
\end{figure}
\cref{fig:failed_cases} illustrates cases where CF-GAP failed to find the search target, meaning that no glimpse location landed on the search target's surface. We observe two main reasons for failures. First, small objects can be surrounded by heavy clutter or be strongly occluded (two examples on the right) so that the glimpsing is diverted towards distracting regions whose texture resembles that of the search target. Second, search targets can be barely distinguishable from their background (two examples on the left).  


\end{document}